\documentclass[a4paper]{svproc}
\usepackage{url}
\usepackage{multicol,multirow}
\usepackage{amsmath,amssymb}
\usepackage{graphicx}
\usepackage{color}
\usepackage[natbib=true]{biblatex}
\graphicspath{{figures/}}

\newcommand{\bo}{\mathbf{o}}

\newcommand{\bu}{\mathbf{u}}

\begin{document}
\mainmatter              
\title{Deep Transfer Learning of Pick Points \\ on Fabric for Robot Bed-Making}
\titlerunning{ISRR 2019 Submission}  
%

\author{Daniel Seita$^{*,1}$, Nawid Jamali$^{*,3}$, Michael Laskey$^{*,4}$, Ron Berenstein$^{1}$, Ajay Kumar Tanwani$^{1,2}$,\\ Prakash Baskaran$^{3}$, Soshi Iba$^{3}$, John Canny$^{1}$, Ken Goldberg$^{1,2}$
\thanks{$^*$ These authors contributed equally.}%
\thanks{$^1$ Department of Electrical Engineering and Computer Sciences; {\scriptsize \{seita,ron.berenstein,ajay.tanwani,canny,goldberg\}@berkeley.edu} }%
\thanks{$^2$ Department of Industrial Engineering and Operations Research}%
\thanks{$^{1-2}$ AUTOLAB at the University of California, Berkeley, USA}
\thanks{$^3$ Honda Research Institute, USA; {\scriptsize\{njamali,pbaskaran,siba\}@honda-ri.com}}%
\thanks{$^4$ Toyota Research Institute, USA; {\scriptsize michael.laskey@tri.global}}%
}

\author{
Daniel Seita\inst{1,}\thanks{These authors contributed equally.} \and
Nawid Jamali\inst{2,*} \and 
Michael Laskey\inst{1,*} \and 
Ajay Kumar Tanwani\inst{1} \and
Ron Berenstein\inst{1} \and
Prakash Baskaran\inst{2} \and
Soshi Iba\inst{2} \and
John Canny\inst{1} \and
Ken Goldberg\inst{1}
}

\authorrunning{Daniel Seita et al.} 
%
%
\institute{University of California, Berkeley; Berkeley, CA 94720, USA.\\
\email{\{seita,laskeymd,ajay.tanwani,ron.berenstein,canny,goldberg\}@berkeley.edu},\\
\and
Honda Research Institute, USA.\\
\email{\{njamali,pbaskaran,siba\}@honda-ri.com}
}

\maketitle              

\begin{abstract}
A fundamental challenge in manipulating fabric for clothes folding and textiles manufacturing is computing ``pick points'' to effectively modify the state of an uncertain manifold. We present a supervised deep transfer learning approach to locate pick points using depth images for invariance to color and texture. We consider the task of bed-making, where a robot sequentially grasps and pulls at pick points to increase blanket coverage. We perform physical experiments with two mobile manipulator robots, the Toyota HSR and the Fetch, and three blankets of different colors and textures. We compare coverage results from (1) human supervision, (2) a baseline of picking at the uppermost blanket point, and (3) learned pick points. On a quarter-scale twin bed, a model trained with combined data from the two robots achieves 92\% blanket coverage compared with 83\% for the baseline and 95\% for human supervisors. The model transfers to two novel blankets and achieves 93\% coverage. Average coverage results of 92\% for 193 beds suggest that transfer-invariant robot pick points on fabric can be effectively learned.
\end{abstract}

\section{Introduction}\label{sec:intro}

Fabric manipulation remains challenging for robots, with real-world applications ranging from folding clothing to handling tissues and gauzes in robotic surgery. In contrast to rigid objects, fabrics have infinite dimensional configuration spaces. In this work, we focus on computing suitable \emph{pick points} for blankets that facilitate smoothing and coverage. 

\begin{figure}[t]
\center
\includegraphics[width=\textwidth]{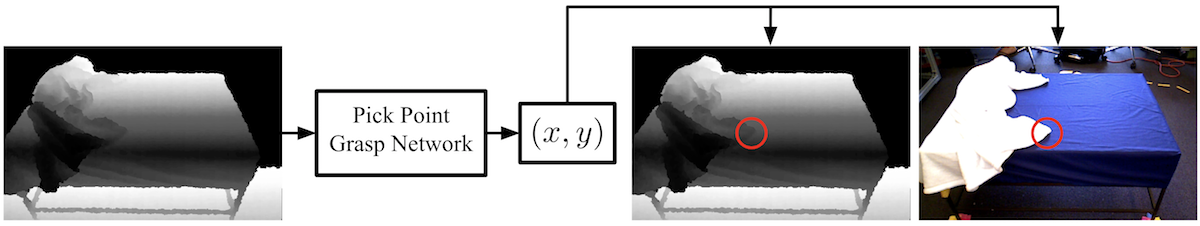}
\caption{System overview: the depth image from the robot's head camera sensors is passed as input to the grasp neural network. The output is the pixel-space pick point $(x,y)$, marked by the red circle in the depth and RGB image pair. The robot grasps at the blanket location corresponding to the pick point, and pulls it towards the nearest uncovered bed surface corner (to the lower right in this example).}
\label{fig:system-overview}
\end{figure}

\begin{figure}[t]
\center
\includegraphics[width=\textwidth]{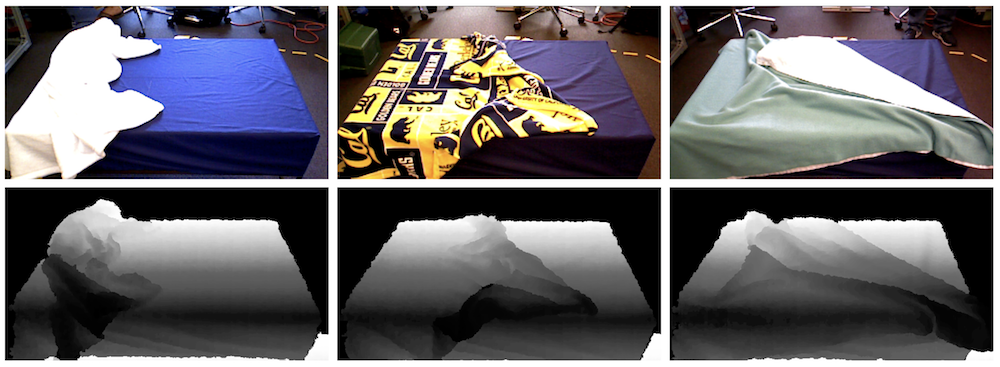}
\caption{
\small
Quarter-scale bed with blue bottom sheet and three different blankets:  white (left), multicolored yellow and blue (Y\&B) pattern (middle), and teal (right). The system is trained on depth images (bottom row).
}
\label{fig:teaser}
\end{figure}

Designing an analytic model is challenging because a blanket is a complex manifold that may be highly deformed, wrinkled, or folded. We present an approach based on deep learning to find pick points from exposed blanket corners, where the blankets are strewn across a flat surface (Figure~\ref{fig:system-overview}). We use depth images as the input modality for invariance to different colors and patterns (Figure~\ref{fig:teaser}).

Consider bed-making with a mobile manipulator. Bed-making is a common home task which is rarely enjoyed and can be physically challenging to senior citizens and people with limited dexterity. Surveys of older adults in the United States~\cite{beer2012domesticated,survey_2009}, suggest that they are willing to have a robot assistant in their homes, particularly for physically demanding tasks. Bed-making is well-suited for home robots since it is tolerant to error and not time-critical~\cite{bloomfield2008recommended,fausset2011challenges}.
While prior work has designed robotic beds equipped with pressure sensors to anticipate patient pose~\cite{2005_robotic_bed} or to lift people in and out of bed~\cite{2010_lift_human}, we apply deep transfer learning to train a single-armed mobile robot with depth sensors to cover a bed with a blanket, without relying on sophisticated bed-related sensors or mechanical features.

The contributions of this paper include: (1) a deep transfer learning approach to selecting pick points that generalizes across robots and blankets, and (2) a formalization of robot bed-making based on pick points and maximizing blanket coverage. We present experimental data from two robots, three blankets, and three pick point methods, demonstrating that learned pick points can achieve coverage comparable to humans. Code and data are available at \url{https://sites.google.com/view/bed-make}.

\section{Related Work}\label{sec:rw}



Fabric manipulation has been explored in a variety of contexts, such as in assistive and home robotics through sewing~\cite{sewing_2012}, ironing~\cite{ironing_2016}, dressing assistance~\cite{2010_assistive_bed,deep_dressing_2018}, and folding of towels and laundry~\cite{laundry2012,shibata2012trajectory}. 

Early research in manipulating fabric used a dual-armed manipulator to hold the fabrics in midair with one gripper, using gravity to help expose borders and corners for the robot's second gripper. Osawa et al.~\cite{osawa_2007} proposed the pick point technique of iteratively re-grasping the lowest hanging point as a subroutine to flatten and classify clothing. Kita et al.~\cite{kita_2009_iros,kita_2009_icra} used a deformable object model to simulate hung clothing, allowing the second gripper to grasp at a desired point.

Follow-up work generalized to manipulating unseen articles of clothing and handling a wider variety of initial cloth configurations. For example, Maitin-Shepard et al.~\cite{maitin2010cloth} identified and tensioned corners to enable a home robot to fold laundry. Their robot grasped the laundry fabric in midair and rotated it to obtain a sequence of images, and used a RANSAC~\cite{ransac} algorithm to fit corners to cloth borders. The robot then gripped any corner with its second gripper, let the cloth settle and hang, and gripped an adjacent corner with its original gripper. Cusumano-Towner et al.~\cite{cusumano2011bringing} followed-up by improving the subroutine of bringing clothing into an arbitrary position. They proposed a hidden Markov model and deformable object simulator along with a pick-point strategy of
regrasping the lowest hanging point. Doumanoglou~et~al.~\cite{unfolding_rf_2014} extended the results by using random forests to learn a garment-specific pick point for folding. These preceding papers rely on gripping the fabrics in midair with a dual-armed robot. In contrast, we focus on finding pick points on fabric strewn across a horizontal surface. Furthermore, the fabrics we use are too large for most dual armed robots to grip while also exposing a fabric corner in midair.

Recently, pick points for fabric manipulation have been learned via reinforcement learning in simulation and then conducting sim-to-real transfer. Thananjeyan et al.~\cite{thananjeyan2017multilateral} developed a tensioning policy for a dual-armed surgical robot to cut gauze. One arm pinched the gauze at a pick point and pulled it slightly. The resulting tension made it easier for the second arm, with a scissor, to cut the gauze. Pick points were selected by uniformly sampling candidates on the gauze and identifying the best one via brute force evaluation. Matas et al.~\cite{sim2real_deform_2018} benchmarked a variety of deep reinforcement learning policies for grasping cloth in simulation, and showed transfer to physical folding and hanging tasks. Their policy outputs a four dimensional action, where the first three are gripper velocities and the last one represents the opening or closing of the gripper for finalizing pick points. Running pure reinforcement learning on physical robots remains difficult due to wear and tear. Thus, these approaches for learning pick points often rely on having accurate environment simulators, but these are challenging to design and not generally available off-the-shelf for robotics tasks.

Additional fabric manipulation techniques include trajectory transfer and learning from image-based wrinkles. Schulman et al.~\cite{schulman_isrr_2013} propose to adapt a demonstrated trajectory (including pick points) from training time to the geometry at test time by computing a smooth, nonrigid transformation. This technique, however, assumes that a nonrigid transformation is possible between two point clouds of the deformable object, which is not generally the case for two different blanket setups. Jia et al.~\cite{cloth_manip_2019} present a cloth manipulation technique which learns from image histograms of wrinkles. While they showed impressive human-robot collaboration results in flattening, folding, and twisting tasks, their dual-armed robot grasped two cloth corners at initialization and kept the grippers closed while moving the arms. Thus, the pick point decision is pre-determined, whereas we learn pick points using deep learning on depth maps.

\section{Methodology}\label{sec:PS}



\subsubsection{Problem Statement} We assume a mobile robot with an arm, a gripper, and color and depth cameras. We assume the robot can position itself to reach anywhere on the half of the bed closest to its position. Let $\pi_\theta : \mathbb{R}^{640\times480} \rightarrow \mathbb{R}^{2}$ be a function parameterized by $\theta$ that maps a depth image $\bo_t \in \mathbb{R}^{640\times 480}$ at time $t$ to a pick point $\bu_t = \pi_\theta(\bo_t)$ for the robot. We are interested in learning the parameters $\theta$ such that the robot grasps at the pick point and pulls the blanket to the uncovered bed frame corner nearest to it.

We represent the resulting blanket configuration with an occupancy function $\xi:\mathbb{R}^3 \rightarrow \lbrace 0, 1 \rbrace$ to determine if a point is part of the blanket or not. Let $c: \xi
\rightarrow \mathbb{R}$ be a function representing a desired performance metric. We define $c(\xi)$ as \emph{blanket coverage}, and measure it from a top-down camera as the percentage of the top bed plane covered by blanket $\xi$. 



\subsection{Setup}\label{sec:setup} Figure~\ref{fig:system-overview} shows an overview of the system. A depth image is presented to a grasp network (described in Section~\ref{ssec:neural_net}) which estimates the location of a suitable pick point. The robot then moves its gripper to the location, closes it, and pulls in the direction of the nearest uncovered corner of the bed frame. Due to the stochastic nature of the task, the robot may not be able to achieve sufficient coverage in a single attempt. For example, the blanket may slip out of the robot's grip during pulling. The robot uses a bed coverage heuristic to decide if it should attempt another grasp and pull at the same side. In this work, we limit the robot to four attempts at each of the two sides of the bed.


\subsection{Grasp Network for Pick Points}\label{ssec:neural_net}

\begin{figure}[t]
\center
\includegraphics[width=0.80\textwidth]{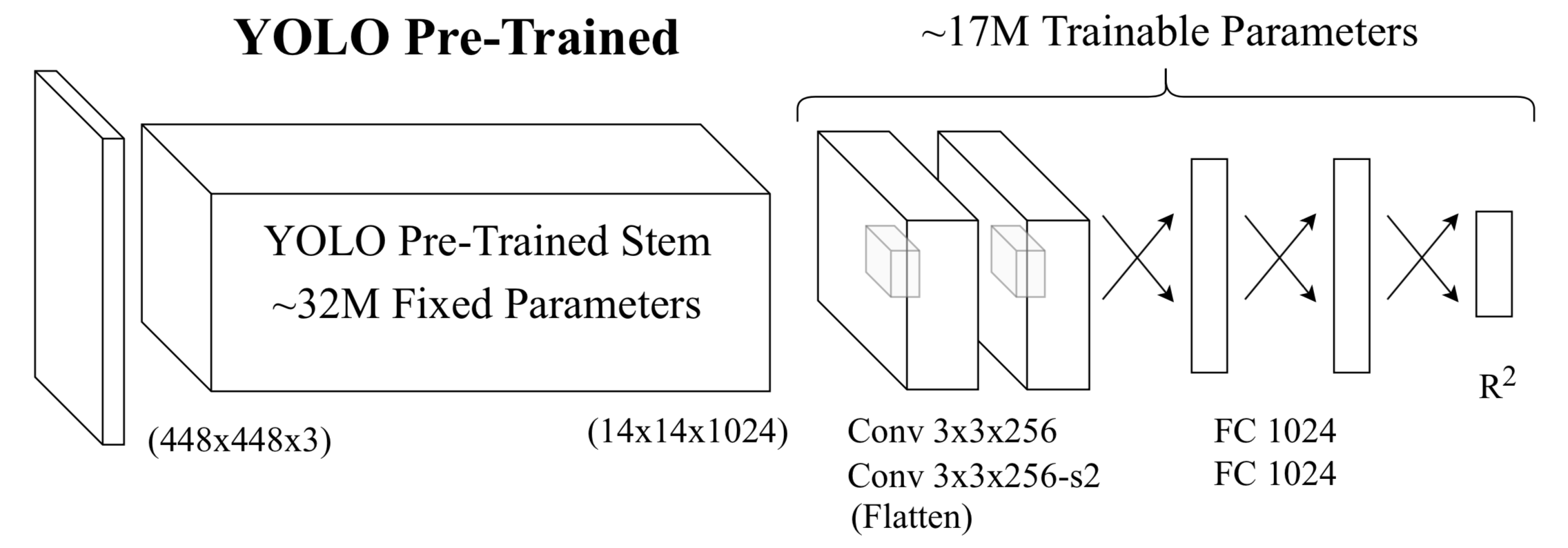}
\caption{
\small
The grasp network architecture. From a $(448\times 448 \times 3)$-sized input image, we use pre-trained weights to obtain a $(14\times 14\times 1024)$ tensor. The input is resized from the original $(640\times 480)$ depth images. We triplicate the depth image along three channels for compatibility with pre-trained weights. The notation: ``-s2'' indicates a stride of two, and two crossing arrows are a dense layer.
}
\vspace*{-5pt}
\label{fig:net_yolo}
\end{figure}

The robot captures the depth image from its head camera sensors as \emph{observation} $\bo_t$. We define a \emph{grasp network} $\pi_{\theta}$ as a deep convolutional neural network~\cite{krizhevsky2012imagenet} that maps from observation $\bo_t$ to a pixel position $\bu_t = (x,y)$ where the robot will grasp (i.e., the pick point). We project this point onto the 3D scene by first measuring the depth value, $z$, from the corresponding depth image. We then project $(x,y,z)$ using known camera parameters, and set the gripper orientation to be orthogonal to the top surface of the bed.

The network $\pi_{\theta}$ is based on YOLO~\cite{redmon2016you}, a single shot object detection network for feature extraction. We utilize pre-trained weights optimized on Pascal VOC 2012~\cite{pascal-voc-2012}. We call this network \emph{YOLO Pre-Trained}, and show its architecture in Figure~\ref{fig:net_yolo}. We fix the first 32 million parameters from YOLO and optimize two additional convolutional layers and two dense layers. 

Since YOLO Pre-Trained has weights trained on RGB images and we use \emph{depth} images with the single channel repeated three times to match RGB dimensions, we additionally tested full training of YOLO without fixing the first 32 million parameters. This, however, converged to a pixel error twice as large.

\section{Data and Experiments}\label{sec:experiments}

\begin{figure}[t]
\center
\includegraphics[width=0.85\textwidth]{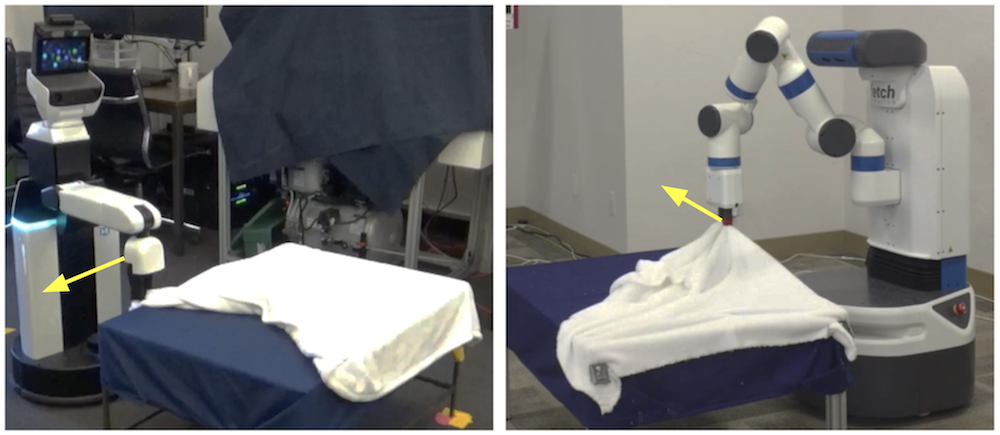}
\caption{
\small
The HSR (left) and Fetch (right) performing the bed-making task with a white blanket. The goal is to cover the (dark blue) top bed surface under the blanket. The yellow arrows indicate the direction of the arm motion. 
}
\label{fig:both_robots}
\end{figure}

We use two mobile manipulator robots, the HSR~\cite{hsr2013} and the Fetch~\cite{fetch} (see Figure~\ref{fig:both_robots}) to evaluate the generality of this approach. The HSR has an omnidirectional base and a 3~DoF arm. The Fetch has a longer 7~DoF arm and a differential drive base. Both robots have PrimeSense head camera sensors with RGB and depth sensors. The robot's task is to make a quarter-scale bed with dimensions $W = 67$ cm, $H = 45$ cm, and $L = 91$ cm. The bed consists of one blanket with area slightly larger than the top surface so that a human can comfortably cover it. One end of the blanket is fixed to one of the shorter sides of the bed frame to simulate two corners being tucked under a mattress. Figure~\ref{fig:both_robots} shows a third-person view of the experimental setup with both robots.

\subsection{Data Collection and Processing}\label{ssec:data_collection}

\begin{figure*}[t]
\center
\includegraphics[width=1.00\textwidth]{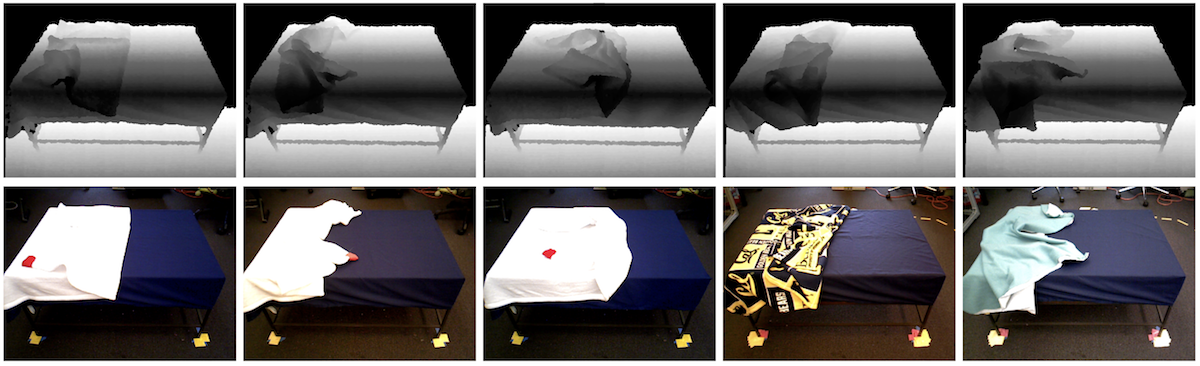}
\caption{
\small
Examples of initial states. The grasp network is trained with depth images (top row). To avoid background noise, we black out regions beyond the validation-tuned depth value of 1.4 meters. We also show the corresponding RGB images (bottom row). The training data is automatically labeled with the red marker from the RGB image. During testing, a network trained on the white blanket is sometimes applied on the Y\&B and the teal blankets shown in the last two columns.
}
\label{fig:init_state_distribution}
\end{figure*}

To facilitate automatic labeling, we used a white blanket with a red mark at its corner to define the pick point. The red mark is solely used for training labels, as the grasp network does not see it in the depth images on which it is trained.

To sample initial states, the human supervisors (the first two authors) fixed a blanket end to one of the shorter edges of the bed and tossed the remaining part onto the top surface. If the nearest blanket corner was not visible or unreachable from the robot's position, the supervisors re-tossed the blanket. Figure~\ref{fig:init_state_distribution} demonstrates examples of initial states as viewed through the robot's head camera sensors. From the initial state, the human supervisors performed short pulls of the blanket and recorded the robot's depth camera image and the pick point after each pull. We took 1028 and 990 images from the camera sensors of the HSR and Fetch, respectively, which we oriented at roughly the same position to keep the viewing angle consistent. The resulting dataset $\mathcal{D} = \{(\bo_i,\bu_i)\}_{i=1}^N$ is used for training the grasp network with $N=2018$. Labels $\bu_i$ correspond to pixel coordinates of the red marked location in $\bo_i$. 

We performed several data pre-processing steps to better condition the optimization. To avoid noise from distant background objects, we set all depth values beyond 1.4 meters from the robot's head camera to zero. Then, depth values are scaled into the range $[0,255]$ to form $\bo_i$, matching the scale of pixels in the RGB images used for pre-trained weights.
We also apply the following data augmentation techniques on $\bo_i$: adding uniform noise in the range $[-4,4]$, adding zero-mean Gaussian noise with standard deviation $\sigma=15$, adding black dots randomly to 0.4\% of the pixels, adding black or white dots randomly (again, to 0.4\% of the pixels), and a flip about the vertical axis to simulate being on the opposite side of the bed. These techniques result in 10x more training data. 

The parameters of $\pi_{\theta}$ are optimized via Adam~\cite{adam2015} by minimizing the $L_2$ loss. The learning rate and $L_2$ regularization strength hyperparameters are chosen based on 10-fold cross-validation performance, where for each fold we train until the validation $L_2$ error stops decreasing. For deployment, we train the grasp network using the combined data from all 10 folds, on the best hyperparameters.

\section{Results}

In this section, we present pick point prediction and blanket coverage results.

\subsection{Grasp Network Training}\label{ssec:dnn_results}

\begin{figure*}[t]
\center
\includegraphics[width=1.00\textwidth]{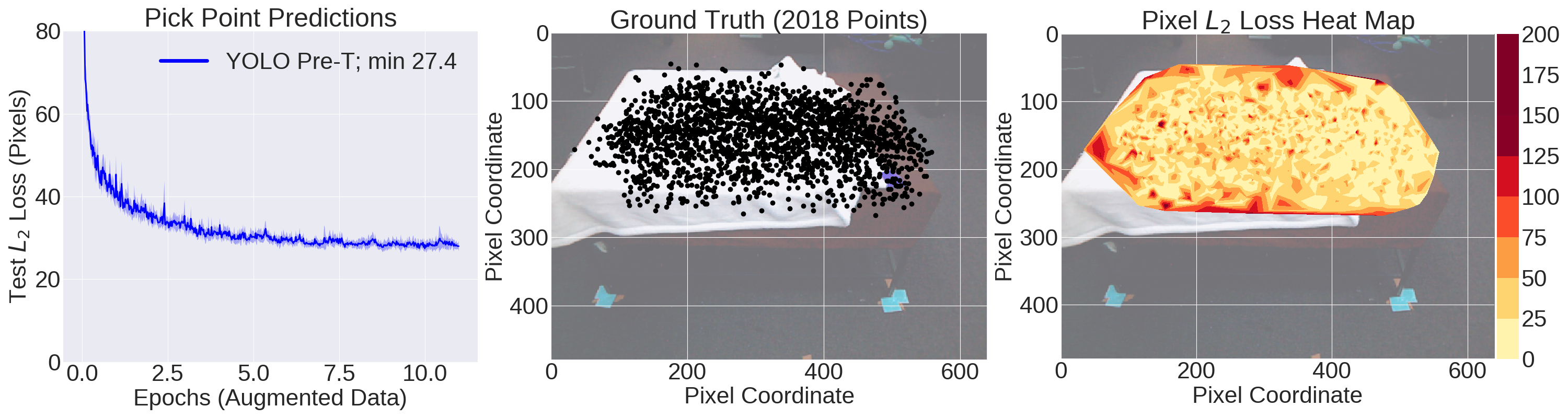}
\caption{
\small
Left: validation set $L_2$ errors (in pixels) when training the grasp network on depth data, averaged over 10-fold cross validation with one standard deviation shaded. Middle: scatter plot showing the distribution of corners (i.e., pick points) in the combined data of 2018 images. Right: heat map of the $L_2$ error in pixel space. The second and third subplots are overlaid on top of a representative image of the bed to aid visualization; see Section~\ref{ssec:dnn_results} for details. We report results for the model with best validation set performance.
}
\label{fig:dnn_results_1}
\end{figure*}

To evaluate the accuracy of pick point estimation, we analyzed the $L_2$ error between the estimated pick point and the ground truth. Figure~\ref{fig:dnn_results_1} (left) demonstrates training results of the grasp network over the best hyperparameter set. It shows the $L_2$ pixel prediction losses as a function of training epoch, indicating that it converges to $27$ pixel error. This pixel error for the grasp network corresponds to the 93\% and 89\% coverage results for the network that we later report in Section~\ref{ssec:coverage_results} and Figure~\ref{fig:results_1}.

Figure~\ref{fig:dnn_results_1} also presents a scatter plot of the distribution of training points (i.e., pick points) and a heat map of those points and their held-out $L_2$ losses in pixels for the best-performing validation set iteration.
Not all the scatter plot points are on top of the bed surface shown in Figure~\ref{fig:dnn_results_1}. The scatter plot and heat map are only overlaid over a \emph{representative} image that the robot might see during the task. Though we set the robot to be at roughly the same position each time we collect data or run the task, in practice there are variations due to imperfections in robotic base motion so the precise pixel location of the bed is not fixed. We observe that the heat map shows darker regions towards the extremes of the dataset, particularly to the left and bottom. These correspond to when blanket corners occur near the edge of the bed surface.


\begin{figure}[t]
\center
\includegraphics[width=0.60\textwidth]{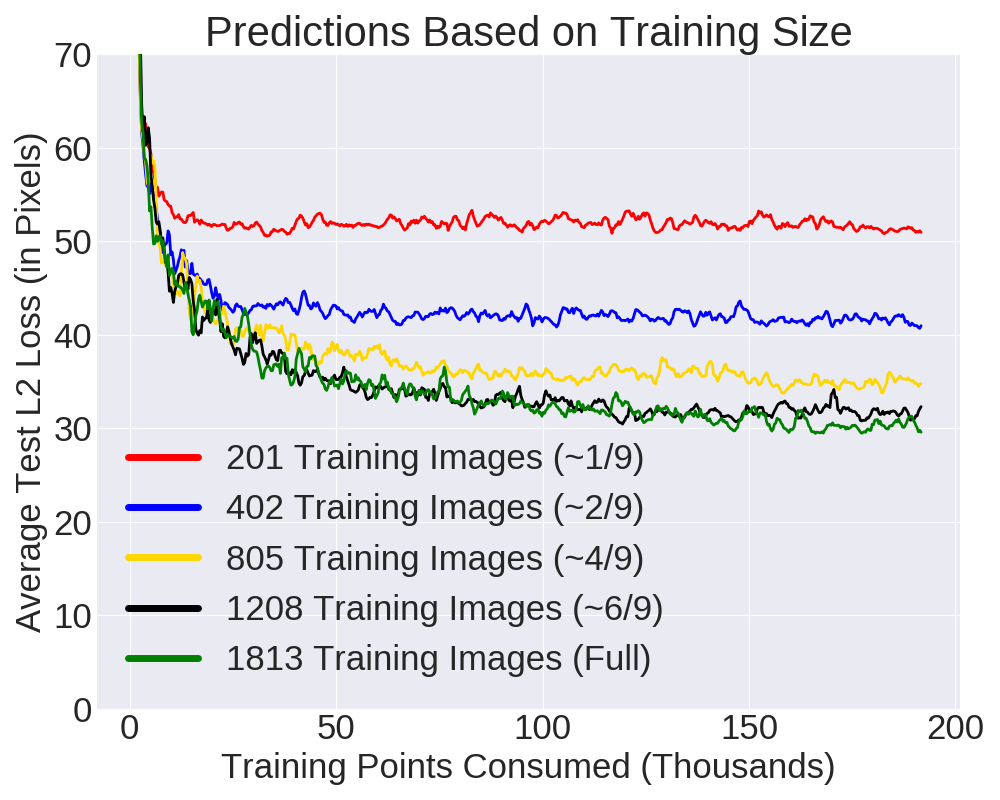}
\caption{
\small
Average $L_2$ pixel error on a held-out set of 205 testing images for five training sets of different sizes. Data sizes range from 201 to 1813 images. Results are a function of the cumulative number of data points consumed (and not epochs, which depend on the overall data size) during training, so the curves also measure efficiency of reducing the $L_2$ pixel error. The results suggest diminishing returns on more data once we hit around 2/3 (i.e., 1208 images) of the largest size here.
}
\label{fig:dnn_ablation_1}
\end{figure}

We next investigate the effect of training data size on $L_2$ pixel error of the pick points. We set aside a held-out set of 205 images, roughly the size of one fold in 10-fold cross validation. We  use the remaining $2018-205 = 1813$ images for training, which we sampled to get 4 training datasets of increasing sizes: $\frac{1}{9}, \frac{2}{9}, \frac{4}{9}, \frac{6}{9}$ of the training dataset. Figure~\ref{fig:dnn_ablation_1} plots a single training run for each of the subsets, as a function of total training points consumed. The number of points consumed is the number of Adam gradient steps multiplied by the batch size, which was fixed at 32. The results suggest a clear benefit to having more training data, with curves corresponding to larger datasets converging to lower $L_2$ error. Nonetheless, we observe diminishing returns at roughly two-thirds of the full training set. While the largest training size gets around 30 $L_2$ error, using about two-thirds of it can attain almost the same test error. For the rest of this paper, we report results with $\pi_\theta$ trained on all the training data.

\subsection{Harris Corner Detector is Insufficient}\label{ssec:harris-heuristics}

\begin{figure}[t]
\center
\includegraphics[width=\textwidth]{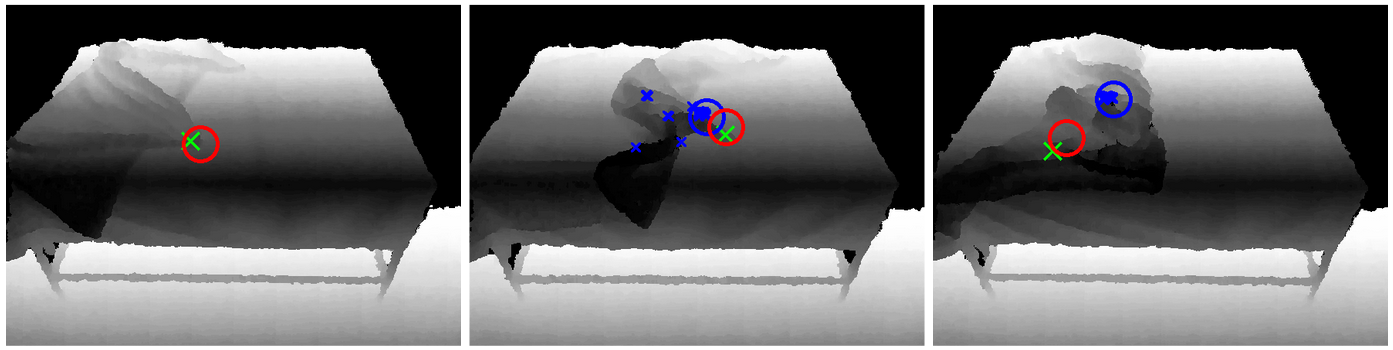}
\caption{Three representative depth images, with blanket corner estimates overlaid from the Harris Corner Detector (blue ``X'') and the grasp network (red circle), along with the ground truth (green ``X''). The blue circle represents the bottom right corner from the set of candidates found by the Harris Corner Detector. Left: no corners are detected on the top surface. Middle: the detector finds many false positives, but some corners are close to the ground truth. Right: all detected corners are false positives. 
}
\label{fig:corner-detector}
\end{figure}

We also investigated whether classical corner detection methods such as the Harris Corner Detector (HCD)~\cite{harris_1988} can be used for selecting blanket corners as pick points. We applied the HCD to get a set of candidates. Then, we selected the corners lying on the bed surface, and picked the one closest to the bottom right of the image. We used a detector tuned for high sensitivity\footnote{We used the OpenCV implementation with block size at 2, Sobel derivative aperture parameter at 1, and free parameter $k$ at 0.001.} on 202 depth images of our dataset (Section~\ref{ssec:data_collection}). Figure~\ref{fig:corner-detector} shows three representative depth images with corner estimates from the grasp network, the HCD, and the ground truth. In some instances HCD fails to detect any corners, and in others, it returns many false positives. Overall, it failed to detect corners in 20 images (10\%) and, when a corner was detected, the average $L_2$ pixel error was 175.0, about 6x worse than the best-performing grasp network presented in Section~\ref{ssec:dnn_results}. Based on these results, we do not pursue the corner detection approach in the rest of this paper.

\subsection{Physical Robot Deployment Evaluation}\label{ssec:coverage_results}

To evaluate the proposed system, we deployed the trained network on two robots. We benchmark the results versus two alternative pick point methods: an analytic highest-point baseline and a human supervisor. For each experimental condition, we report average coverage before and after each rollout. To evaluate blanket coverage, we measure the area of the bed's top surface and the area of its uncovered portion using contour detection on a top-down camera image. All images used for coverage results in the following experiments are on the project website.

\subsubsection{Bed-Making Rollout}

To start each rollout, the robot is positioned at one of the longer sides of the bed. The robot determines a pick point on the blanket based on one of three methods (analytic, human, or learned) then grasps and pulls it to the nearest uncovered bed corner. The robot repeats the process on the second side of the bed.

As stated in Section~\ref{sec:setup}, the robot is allowed up to four grasp and pull attempts per side. Thus, after each attempt, the robot visually checks if its action resulted in sufficient coverage. We can use a variety of heuristics to measure coverage and encode this behavior. For these experiments, we trained a second deep neural network with the same architecture as the grasp network. We trained it to detect, given the depth image, if the side closest to the robot is sufficiently covered. This network was applied to all experimental conditions to allow us to focus on comparing pick point methods.


\subsubsection{Initialization and Pick Point Method Selection}

To ensure the evaluations are fair with respect to the initial blanket state, the blanket is tossed on the the bed frame, \emph{then} one of the three pick point methods (analytic, human, or learned) is selected at random. This was repeated until we achieved a minimum of 24 and 15 rollouts per method for the HSR and Fetch, respectively.

\subsubsection{Analytic Highest Point Baseline}\label{sssec:method1}

\begin{figure*}[t]
\center
\includegraphics[width=0.9\textwidth]{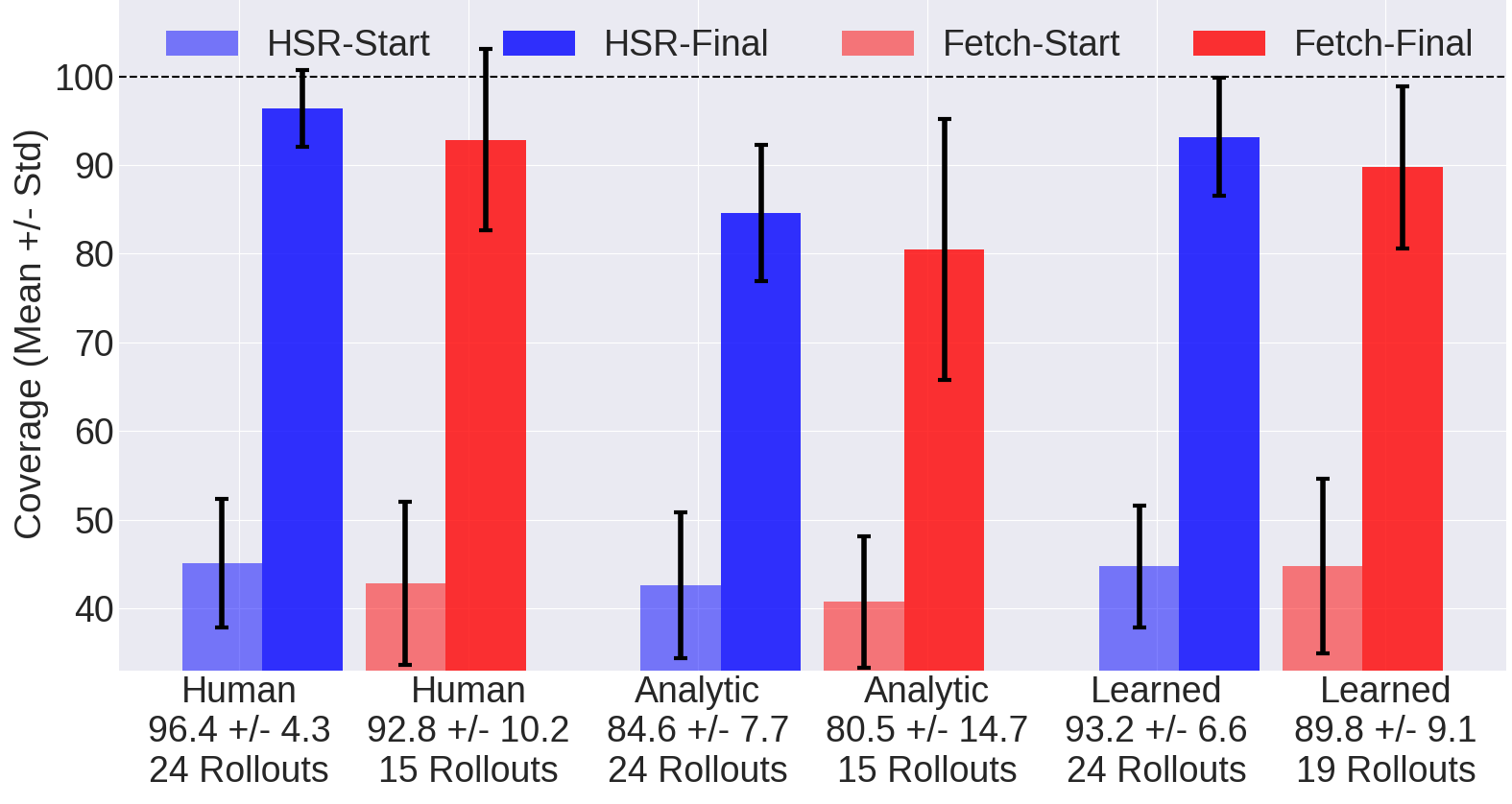}
\caption{
\small
Bed coverage results with the HSR (blue) and Fetch (red) at the start and end of each rollout, all applied on a bed with the white blanket. \emph{Human}: human selecting pick points via a click interface. \emph{Analytic}: the highest reachable point baseline. \emph{Learned}: depth-based neural network, the pick point method we propose in this paper based on depth images of the blanket. The error bars represent one standard deviation.
}
\label{fig:results_1}
\end{figure*}


We benchmark with an analytic baseline where the robot grips the highest reachable point on the blanket. Figure~\ref{fig:results_1} shows that the analytic baseline achieves $85\pm 8 \%$ and $81\pm 15 \%$ coverage for the HSR and Fetch, over 24 and 15 rollouts, respectively.

\begin{figure}[t]
\center
\includegraphics[width=0.85\textwidth]{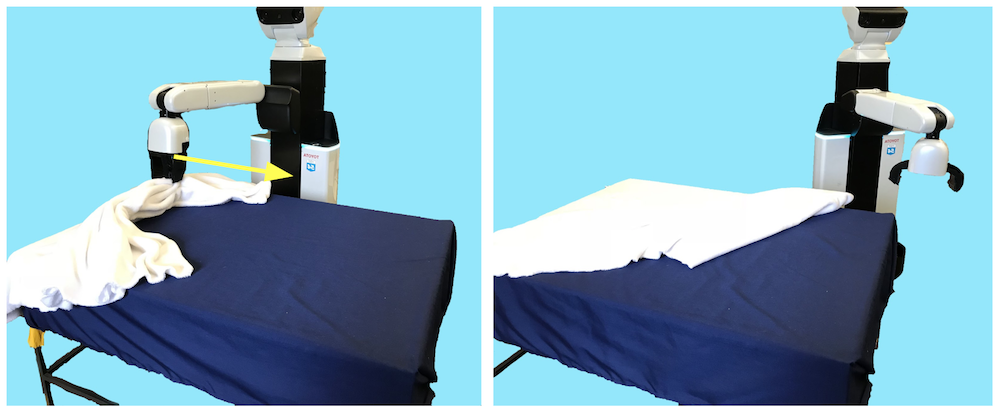}
\caption{
\small
The HSR following the analytic baseline. It grasps at the highest reachable point and pulls in the direction of the yellow arrow, improving coverage.
}
\label{fig:failure_1}
\end{figure}

The analytic method performs reasonably well, but has high variance. The highest point may correspond to a corner fold, in which case the analytic method will significantly increase coverage. When the pick point is not at a corner fold, the analytic method tends to use multiple grasp and pull actions, which also increases coverage. After the first attempt, however, the blanket corner frequently gets folded under, limiting future coverage increases. Figure~\ref{fig:failure_1} shows the HSR following the highest point baseline grasping procedure. After the robot grasps and pulls the highest point, it attains significant coverage, but the blanket is at a state that little further improvements can be made.

\subsubsection{Human Supervisor}\label{sssec:method2}

As another benchmark, we have humans (the first two authors) select pick points via a click interface over a web-server application.  The human was only provided the usual depth image as input, to be consistent with the input to the grasp network, and did not physically touch the robot or the blanket. Figure~\ref{fig:results_1} suggests that the human supervisor achieves $96\pm 4 \%$ and $93\pm 10 \%$ coverage for the HSR and Fetch, over 24 and 15 trials, respectively.

\subsubsection{Learned Grasp Network}\label{sssec:method3}

\begin{figure*}[t]
\center
\includegraphics[width=0.65\textwidth]{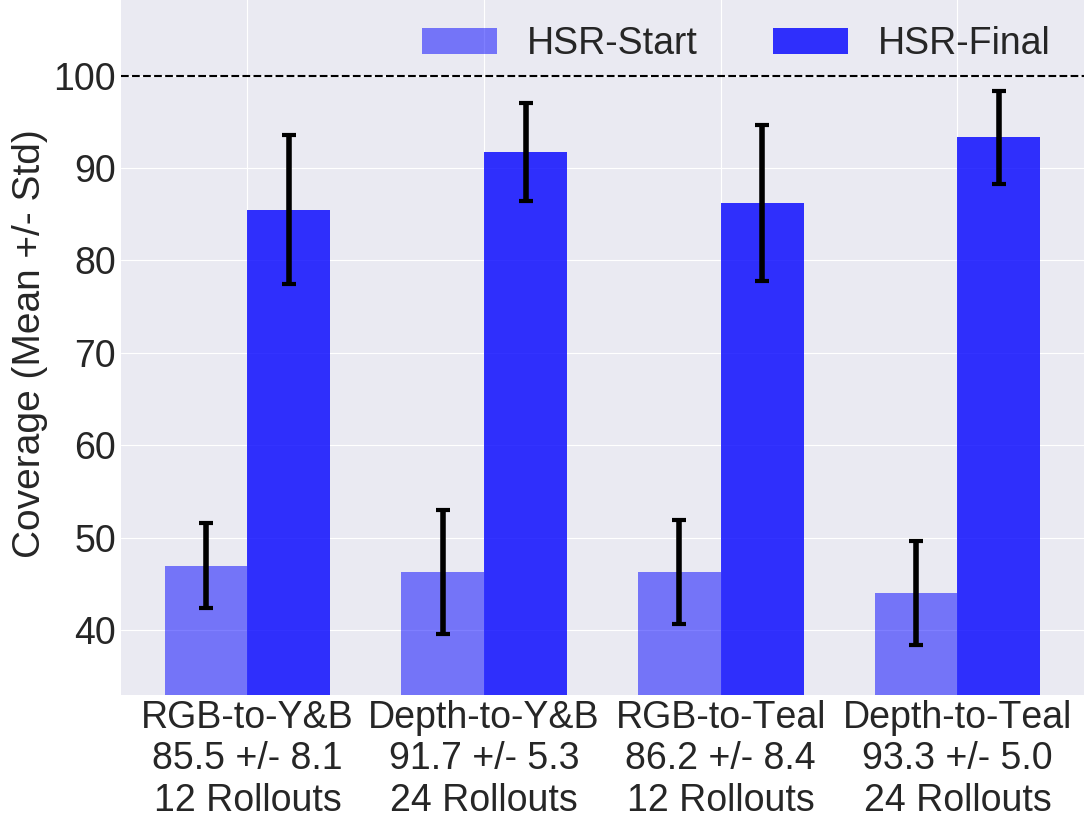}
\caption{
\small
Results of generalization to blankets with different colors and patterns using the HSR. \emph{RGB-to-Y\&B} and \emph{RGB-to-Teal}: the RGB-based grasp network on Y\&B and teal blankets. \emph{Depth-to-Y\&B} and \emph{Depth-to-Teal}: depth-based grasp network on Y\&B and teal blankets. The error bars represent one standard deviation.
}
\label{fig:results_generalization}
\end{figure*}

Figure~\ref{fig:results_1} shows that on the white blanket, the learned grasp network $\pi_\theta$ attains $93\pm 7 \%$ and $90\pm 9 \%$ coverage for the HSR and Fetch, over 24 and 19 trials, respectively (with an average coverage of 92\% for both robots combined). This outperforms the analytic baseline, demonstrating the feasibility of a learning based approach. Moreover, the trained grasp network performs nearly as well as the human supervisor. These results are consistent among the two robots, providing evidence of the robot-to-robot transfer capability of the method. The accompanying video shows a rollout of the learned grasp network deployed on both robots.


To test for statistical significance among different experimental conditions, we use the Mann-Whitney U test~\cite{mann_whitney}. The standard $t$-test is not used as the coverage metric is not normally distributed. For the HSR, the Mann-Whitney U test for the analytic versus learned network (when applied on the white blanket setup) is $p=0.00034$, a strong indicator of statistical significance. The same test for the learned network versus a \emph{human} results in a higher value of $p=0.0889$, suggesting that the learned grasp network's performance more closely matches the human supervisor than the analytic baseline.

\subsubsection{Average Number of Grasps}

We analyzed the average number of grasp attempts made by each method. Each rollout was limited to a maximum of eight grasp attempts (four per side). The highest point baseline required an average of 6.2 and 4.9 attempts for the HSR and Fetch, respectively, compared to 4.4 and 4.3 for the learned network, and 2.8 and 3.0 for the human supervisor. The baseline method used more attempts because it often covered the blanket corner during the first pull and failed to make progress in the remaining grasps before reaching the limit of four attempts per side.

\subsubsection{Depth-Based Grasp Network Generalization}

We tested the generalization capability of the method to blankets with different colors and patterns: the multicolored Y\&B and teal blankets shown in Figure~\ref{fig:teaser}. We deployed the same grasp network (trained on depth images of the white blankets) on the HSR with 24 rollouts for each of the two other blankets. The results in Figure~\ref{fig:results_generalization} show that the HSR attains $92\pm 5 \%$ and $93\pm 5 \%$ coverage for the Y\&B and teal blankets, respectively. The Mann-Whitney U tests for the depth-based grasp network on white versus Y\&B, white versus teal, and teal vs Y\&B blankets (24 rollouts for each comparison) are $p=0.227$, $p=0.844$, and $0.327$, respectively. These relatively high $p$-values mean we cannot reject a hypothesis that the coverage samples in each group are from the same distribution, suggesting that the grasp neural network trained on depth images (of the white blanket) directly transfers to two other blankets despite slightly different material properties; the Y\&B blanket is thinner, while the teal blanket is less elastic and has a thin white sheet pinned underneath it.

\subsubsection{RGB-Based Grasp Network Generalization}

We trained a new grasp network using the \emph{RGB images} of the white blanket to compare with the depth-based approach. Other than this change, the RGB-based network was trained in an identical manner as the depth-based network. From Figure~\ref{fig:results_generalization}, we observe that the RGB-based network obtains $86\pm 8 \%$ and $86\pm 8 \%$ coverage on the Y\&B and teal blankets, suggesting that the depth-based method generalizes better to different colors and patterns. Empirically, we observe that the RGB-based network consistently grasps the Y\&B blanket close to the center of its area over the bed's top surface. For the teal blanket, it tends to pick anywhere along the exposed white underside, or near the center if there is no white visible.

\subsection{Timing Analysis}\label{ssec:bed_results_1}

\begin{table}[t]
\caption{
\small
Timing results of bed-making rollouts for the HSR and Fetch, all in seconds, and the number of times the statistic was recorded (``quantity''). \emph{Moving to a Side}: moving from one side of the bed to another, which happens once per rollout. \emph{Grasp Execution}: the process of the robot moving its end-effector to the workspace and pulling to a target. \emph{Neural Network Pass}: the forward pass through the grasp network, which is not recorded for the analytic and human pick point methods.
}
\centering
\begin{tabular}{l l r r}
& \textbf{Component} & \textbf{Mean Time (Sec.)} & \textbf{Quantity} \\ \hline
\multirow{3}{*}{HSR}& Moving to a Side    & $32\pm 2$  & 144 \\
& Grasp Execution     & $18\pm 2$  & 706 \\
& Neural Network Pass & $ 0.1\pm 0.2$  & 491 \\ \hline
\multirow{3}{*}{Fetch}& Moving to a Side    & $29\pm 22$ & 49  \\
& Grasp Execution     & $88\pm 19$ & 201 \\
& Neural Network Pass & $ 0.1\pm 0.2$  & 82  \\
\end{tabular}
\label{tab:timing}
\end{table}

We performed experiments on a single workstation with an NVIDIA Titan Xp GPU, and list timing results in Table~\ref{tab:timing} for the three major components of the bed-making rollout: moving to another side of the bed, physical grasp execution, and grasp network forward passes (if applicable). The reported numbers combine all relevant trials from Figures~\ref{fig:results_1} and~\ref{fig:results_generalization}, with 144 and 49 total rollouts for the HSR and Fetch, respectively. The major bottlenecks are moving to another side, which required $32\pm 2$ s and $29\pm 22$ s for the HSR and Fetch, respectively, and grasp execution, which took $18\pm 2$ s and $88\pm 19$ s. In contrast, the fast single-shot CNN required a mean time of just $0.1\pm 0.02$ s.

\section{Conclusion and Future Work}\label{sec:conclusions}

We presented a supervised learning approach to select effective pick points on fabrics. The method uses depth images to be invariant to the color and pattern of the fabric. We applied the method to a quarter-scale bed-making task, where the robot grasps at a pick point and pulls the blanket to the edge of the bed to increase coverage. We trained a grasp network to estimate pick points, and deployed it on the HSR and Fetch. Results suggest that the proposed depth-based grasp network outperforms an analytic baseline that selects the highest blanket point, and better generalizes to different fabrics as compared to an RGB-based network.

In future work, we will explore scalable representations for learning pick points with Fog robotics~\cite{Tanwani_2019}. We will also extend the bed-making task by reducing the number of hard-coded robot actions and learning policies that can handle a wider variety of blanket configurations. Such policies may be learned using reinforcement learning in combination with cloth simulators to decrease the number of real-world examples required. Finally, we plan to explore application to manipulating furniture covers, table cloths, textiles, and other deformable objects.


\section*{Acknowledgments}
\small
This research was performed at the AUTOLAB at UC Berkeley in affiliation with the Berkeley AI Research (BAIR) Lab, Berkeley Deep Drive (BDD), the Real-Time Intelligent Secure Execution (RISE) Lab, and the CITRIS ``People and Robots'' (CPAR) Initiative, and by the Scalable Collaborative Human-Robot Learning (SCHooL) Project, NSF National Robotics Initiative Award 1734633. The authors were supported in part by donations from Honda Research Institute Siemens, Google, Amazon Robotics, Toyota Research Institute, Autodesk, ABB, Samsung, Knapp, Loccioni, Intel, Comcast, Cisco, Hewlett-Packard and by equipment grants from PhotoNeo, NVidia, and Intuitive Surgical. We thank Ashwin Balakrishna, David Chan, Carolyn Chen, Zisu Dong, Jeff Ichnowski, Roshan Rao, and Brijen Thananjeyan for helpful feedback. Daniel Seita is supported by a National Physical Science Consortium Fellowship.

\printbibliography
\end{document}